\documentclass[11pt,a4paper]{article}
\usepackage[hyperref]{emnlp2020}
\usepackage{times}
\usepackage{latexsym}

\usepackage{graphicx}
\usepackage{todonotes}
\usepackage{mathtools, nccmath}
\usepackage{url}
\usepackage{amsfonts}
\usepackage{array}
\usepackage{subcaption}
\usepackage{makecell}
\usepackage{amssymb}
\usepackage{pifont}
\usepackage{adjustbox}
\usepackage{microtype}

\aclfinalcopy 

\newcommand{\cmark}{\ding{51}}%
\newcommand{\xmark}{\ding{55}}%

\makeatletter
\newcommand\newtag[2]{#1\def\@currentlabel{#1}\label{#2}}
\makeatother

\newcommand{\SideNote}[2]{} 
\renewcommand{\SideNote}[2]{\todo[color=#1,size=\footnotesize]{#2}}

\title{Context-aware Stand-alone Neural Spelling Correction}

\author{Xiangci Li \thanks{\ \  Work performed during internship with Baidu USA} \\
  University of Texas at Dallas \\
  Richardson, TX \\
  \texttt{lixiangci8@gmail.com} \\\And
  Hairong Liu \\
  Baidu USA \\
  Sunnyvale, CA \\
  \texttt{liuhairong@baidu.com} \\\And
  Liang Huang \\
  Baidu USA \\
  Sunnyvale, CA \\
  \texttt{lianghuang@baidu.com} \\}

\date{}

\begin{document}
\maketitle
\begin{abstract}
Existing natural language processing systems are vulnerable to noisy inputs resulting from misspellings. 
On the contrary, humans can easily infer the corresponding correct words 
from their misspellings and surrounding context. Inspired by this, we address the \emph{stand-alone} spelling correction problem, which 
only corrects the spelling of each token without additional token insertion or deletion, by utilizing both spelling information and global context representations. We present a simple yet powerful solution that jointly detects and corrects misspellings as a sequence labeling task by fine-turning a pre-trained language model. Our solution outperform the previous state-of-the-art result by $12.8\%$ absolute $F_{0.5}$ score.
\end{abstract}

\section{Introduction}
A spelling corrector is an important and ubiquitous pre-processing tool in a wide range of applications, such as word processors, search engines and machine translation systems. 
Having a surprisingly robust language processing system to denoise the scrambled spellings, humans can relatively easily solve spelling correction \cite{rawlinson1976significance}.
However, spelling correction is a challenging task for a machine, because words can be misspelled in various ways, and a machine has difficulties in fully utilizing the contextual information. 

Misspellings can be categorized into \emph{non-word}, which is out-of-vocabulary, and the opposite, \emph{real-word} misspellings \cite{klabunde2002daniel}. The dictionary look-up method can detect non-word misspellings, while real-word spelling errors are harder to detect, since these misspellings are in the vocabulary \cite{mays1991context,wilcox2008real}. In this work, we address the \emph{stand-alone} \cite{li2018spelling} spelling correction problem. It only corrects the spelling of each token without introducing new tokens or deleting tokens, so that the original information is maximally preserved for the down-stream tasks. 

We formulate the \emph{stand-alone} spelling correction as a sequence labeling task and jointly detect and correct misspellings. Inspired by the human language processing system, we propose a novel solution on the following aspects:
(1) We encode both spelling information and global context information in the neural network.
(2) We enhance the real-word correction performance by initializing the model from a pre-trained language model (LM).
(3) We strengthen the model's robustness on unseen non-word misspellings by augmenting the training dataset with a synthetic character-level noise. As a result, our best model \footnote{\url{https://github.com/jacklxc/StandAloneSpellingCorrection}} outperforms the previous state-of-the-art result \cite{wang2019learning} by $12.8\%$ absolute $F_{0.5}$ score.

\section{Approach}

\begin{figure}[t]
  \centering
    \includegraphics[width=0.48\textwidth]{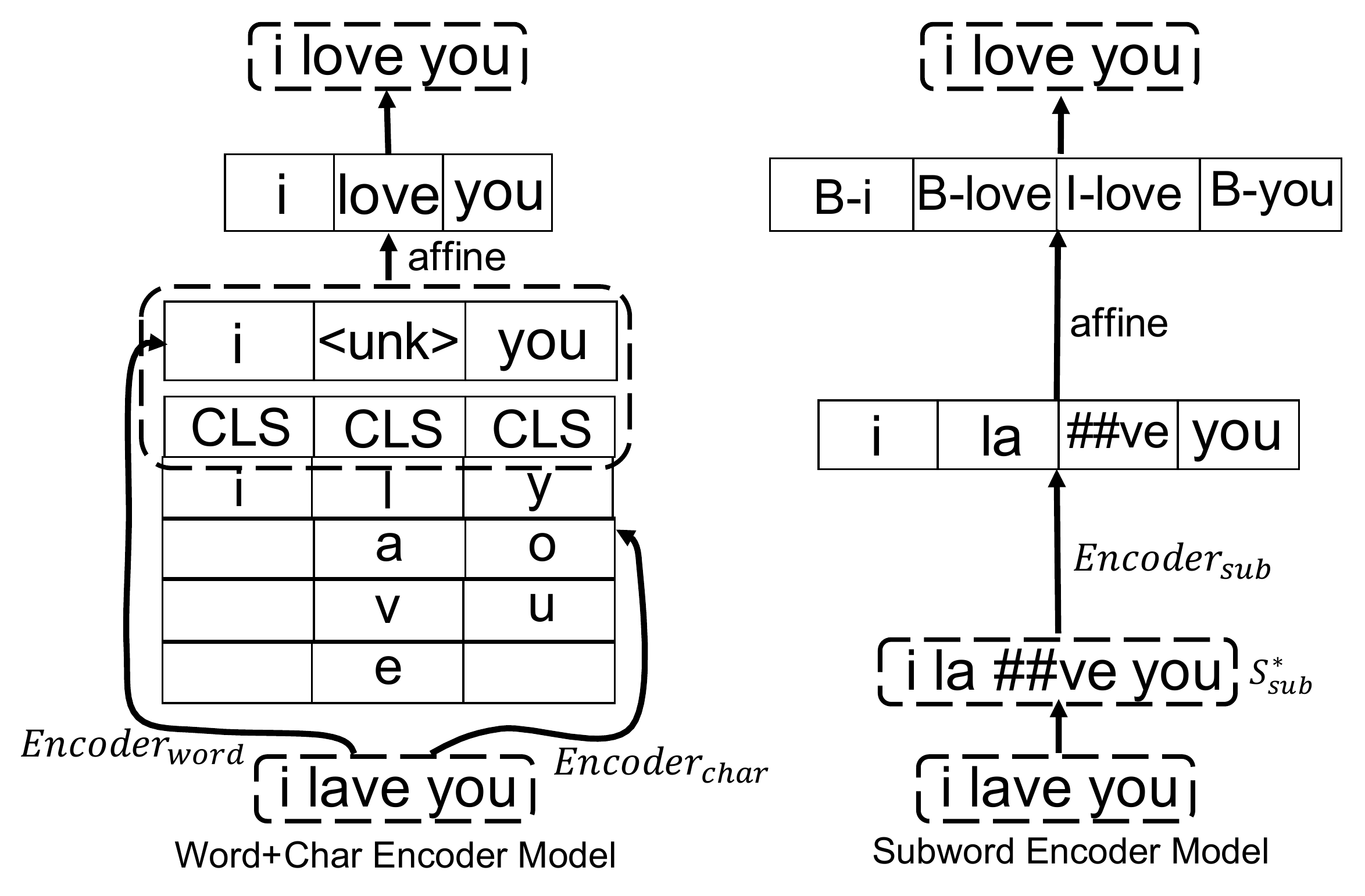}
    \vspace{-1.5em}
    \caption{A schematic illustration of our approach. Left: combined word-level and character-level encoder model. Right: subword-level model using $BIO2$ tagging scheme \cite{sang1999representing}. 
    } \label{fig:spelling_corrector}
    \vspace{-1em}
\end{figure}

We use the transformer-encoder \cite{vaswani2017attention} to encode the input sequences and denote it as $Encoder$. As illustrated in Figure \ref{fig:spelling_corrector}, we present both Word+Char encoder and Subword encoder, because we believe the former is better in encoding spelling information, while the latter has the benefit of utilizing a large pre-trained LM.

\paragraph{Word+Char encoder.}
We use a word encoder to extract global context information and a character encoder to encode spelling information. As shown in equation \ref{eq:word_char}, in order to denoise the noisy word sequence $S^*$ to the clean sequence $S$, we first separately encode $S^*$ using a word-level transformer-encoder $Encoder_{word}$ and each noisy spelling sequence $C^*_{k}$ of token $k$ via a character-level transformer-encoder $Encoder_{char}$. For $Encoder_{word}$, we replace non-word misspellings, i.e. OOV words, with a $\langle unk \rangle$ token. For $Encoder_{char}$, we treat each character as a token and each word as a ``sentence'', so each word's character sequence embedding $h^k_{char}$ is independent of each other. Since the transformer-encoder \cite{vaswani2017attention} computes contextualized token representations, we take $h_{char}$, the $[CLS]$ token representation of each character sequence as the local character-level representation of $S^*$. Finally, we jointly predict $S$ by concatenating the local and global context representations.

\begin{equation}
\small
\begin{aligned}
h_{word} &= Encoder_{word}(S^*) \\
h^k_{char} &= Encoder_{char}(C^*_{k}) \\
h_{char} &= [CLS(h^1_{char}),CLS(h^2_{char}),...,CLS(h^n_{char})]\\
h_{S} &= [h_{word};h_{char}] \\
p(S) &= softmax(Wh_{S}+b))
\end{aligned}
\label{eq:word_char}
\end{equation}


 
\paragraph{Subword encoder.}
Alternatively, we use subword tokenization to simultaneously address the spelling and context information. Formally, as shown in equation \ref{eq:subword}, given a noisy subword token sequence $S^*_{sub}$, we encode it using a transformer-encoder $Encoder_{sub}$ and simply use an affine layer to predict the sequence of each subword token's corresponding correct word token $S_{sub}$ in $BIO2$ tagging scheme \cite{sang1999representing}.

\begin{equation}
\small
\begin{aligned}
h_{sub} &= Encoder_{sub}(S^*_{sub}) \\
p(S_{sub}) &= softmax(W_{sub}h_{sub}+b_{sub})
\end{aligned}
\label{eq:subword}
\end{equation}
Furthermore, we fine-tune our Subword encoder model with a pre-trained LM initialization to enhance the real-word misspelling correction performance.

We use cross-entropy loss as our training objective. Finally, in addition to the natural misspelling noise, we apply a synthetic character-level noise to the training set to enhance the model's robustness to unseen misspelling patterns. The details will be introduced in section \ref{sec:dataset}.

\begin{table*}[h]
\begin{center}
\begin{tabular}{m{0.5em}|m{13em}|m{2em}m{2em}m{2em}m{2.5em}|m{2em}m{2em}m{2em}m{2em}}
 \hline
&Models & \multicolumn{4}{c}{Dev} &  \multicolumn{4}{c}{Test}\\ 
& & Acc & P & R & $F_{0.5}$ & Acc & P & R & $F_{0.5}$ \\ \hline
\newtag{1}{model:ScRNN} & ScRNN \cite{sakaguchi2017robsut}&$0.958$&$0.823$ &$0.890$ &$0.836$&$0.946$ &$0.755$ &$0.865$ &$0.775$  \\
\newtag{2}{model:MUDE} & MUDE \cite{wang2019learning} &$0.966$ &$0.829$ &$0.952$ &$0.851$ &$0.952$ &$0.751$ &$0.928$ &$0.781$  \\  \hline \hline
\newtag{3}{model:char} & Char Encoder &$0.883$ &$0.517$ &$0.819$ &$0.559$ &$0.870$ &$0.458$ &$0.802$ &$0.501$ \\
\hline
\newtag{4}{model:word} & Word Encoder &$0.932$ &$0.565$ &$0.949$ &$0.615$ &$0.924$ &$0.521$ &$0.903$ &$0.570$ \\ \hline
\newtag{5}{model:word_char} & Word + Char Encoder &$0.988$ &$\mathbf{0.959}$ &$0.959$ &$\mathbf{0.959}$ &$0.974$ &$0.882$ &$0.929$ &$0.891$ \\
\newtag{6}{model:word_char_randchar} & + random char &$0.986$ &$0.953$ &$0.947$ &$0.951$ &$0.976$ &$\mathbf{0.898}$ &$0.927$ &$0.904$ \\
\hline
\newtag{7}{model:wordpiece} & Subword Encoder &$0.986$ &$0.934$ &$0.972$ &$0.941$ &$0.968$ &$0.831$ &$0.950$ &$0.852$  \\
\newtag{8}{model:wordpiece_char} & + Char Encoder &$0.980$ &$0.908$ &$0.959$ &$0.917$ &$0.963$ &$0.808$ &$0.939$ &$0.831$  \\
\newtag{9}{model:wordpiece_randchar} & + random char &$0.985$ &$0.931$ &$0.966$ &$0.938$ &$0.973$ &$0.866$ &$0.950$ &$0.881$  \\
\newtag{10}{model:ernie} & + LM pre-train &$\mathbf{0.990}$ &$0.951$ &$\mathbf{0.982}$ &$0.957$ &$0.975$ &$0.866$ &$0.962$ &$0.883$  \\
\newtag{11}{model:ernie_randchar} & + LM pre-train + random char &$0.989$ &$0.946$ &$0.979$ &$0.952$ &$\mathbf{0.980}$ &$0.896$ &$\mathbf{0.964}$ &$\mathbf{0.909}$  \\
 \hline
 \hline

\end{tabular}
 \caption{Model performance and ablation studies measured by accuracy, precision, recall and $F_{0.5}$.}
   \label{tab:result}
 \end{center}
\end{table*}

\begin{table*}[h]
\begin{center}
\begin{tabular}{m{0.5em}|m{13em}|m{2em}m{2em}m{2.2em}|m{2em}m{2em}m{2.2em}||m{2.2em}|m{2em}}
 \hline
&Models & \multicolumn{6}{c}{Real-Word} &  \multicolumn{2}{c}{Non-Word}\\ 
& & \multicolumn{3}{c}{dev} & \multicolumn{3}{c}{test} & \multicolumn{1}{c}{dev} &\multicolumn{1}{c}{test}\\ 
& & P & R & $F_{0.5}$ & P & R & $F_{0.5}$ & P & P\\ \hline
\newtag{1}{model:ScRNN} & ScRNN \cite{sakaguchi2017robsut} &$0.507$ &$0.592$ &$0.522$ &$0.456$ &$0.523$ &$0.468$ &$0.952$ &$0.873$  \\
\newtag{2}{model:MUDE} & MUDE \cite{wang2019learning} &$0.595$&$0.825$&$0.630$&$0.533$&$0.747$ &$0.566$ &$0.945$ &$0.855$ \\
 \hline \hline
\newtag{3}{model:char} & Char Encoder &$0.106$ &$0.304$ &$0.122$ & $0.099$ &$0.296$ &$0.113$ &$0.886$ &$0.792$ \\
\hline
\newtag{4}{model:word} & Word Encoder &$\mathbf{0.916}$ &$0.889$ &$\mathbf{0.911}$ &$\mathbf{0.835}$ &$0.792$ &$\mathbf{0.826}$ &$0.438$ &$0.414$ \\
\hline
\newtag{5}{model:word_char} & Word + Char Encoder &$0.900$ &$0.851$ &$0.900$ &$0.819$ &$0.750$ &$0.804$ &$0.979$ &$0.903$ \\
\newtag{6}{model:word_char_randchar} & + random char &$0.902$ &$0.807$ &$0.881$ &$0.819$ &$0.741$ &$0.802$ &$0.969$ &$0.924$ \\
\hline
\newtag{7}{model:wordpiece} & Subword Encoder &$0.804$ &$0.897$ &$0.821$ &$0.715$ &$0.827$ &$0.735$ &$\mathbf{0.988}$ &$0.877$ \\
\newtag{8}{model:wordpiece_char} & + Char Encoder &$0.740$ &$0.848$ &$0.759$ &$0.664$ &$0.786$ &$0.685$ &$0.978$ &$0.867$ \\
\newtag{9}{model:wordpiece_randchar} & + random char &$0.799$ &$0.876$ &$0.813$ &$0.718$ &$0.819$ &$0.736$ &$0.984$ &$0.925$ \\
\newtag{10}{model:ernie} & + LM pre-train &$0.850$ &$\mathbf{0.935}$ &$0.866$ &$0.771$ &$0.870$ &$0.789$ &$\mathbf{0.988}$ &$0.877$ \\
\newtag{11}{model:ernie_randchar} & + LM pre-train + random char &$0.845$ &$0.922$ &$0.860$ &$0.787$ &$\mathbf{0.872}$ &$0.803$ &$0.987$ &$\mathbf{0.941}$ \\
 \hline
 \hline
\end{tabular}
 \caption{Real-word and non-word performance measured by precision, recall and $F_{0.5}$. \textbf{All of the recall of non-word is 1.000.}}
   \label{tab:detailed_result}
 \end{center}
\end{table*}

\section{Experiments}
\subsection{Dataset} \label{sec:dataset}
Since we cannot find a sentence-level misspelling dataset, we create one by using the sentences in the $1$-Billion-Word-Language-Model-Benchmark \cite{chelba2013one} as gold sentences and randomly replacing words with misspellings from a word-level natural misspelling list \cite{mitton1985corpora,belinkov2017synthetic} to generate noisy input sentences. In a real scenario, there will always be unseen misspellings after the model deployment, regardless of the size of the misspelling list used for training. Therefore, we only use $80\%$ of our full word-level misspelling list for \emph{train} and \emph{dev} set. In order to strengthen the robustness of the model to various noisy spellings, we also add noise from a character-level synthetic misspelling list \cite{belinkov2017synthetic} to the training set. As a result, real-word misspelling contributes to approximately $28\%$ of the total misspellings for both \emph{dev} and \emph{test} set. The details are described in Section \ref{sec:dataset_detail}

\subsection{Results}
\label{sec:results}


\paragraph{Performance Metrics}
\label{sec:performance_metrics}
\begin{table}[h]
\begin{center}
\begin{tabular}{c|c|c} 
 \hline
$=$ Ground Truth? & Noisy Input & Prediction \\ \hline
True Positive & \xmark & \cmark \\ \hline
False Positive & \cmark & \xmark \\ \hline
False Negative & \xmark & \xmark \\ \hline
True Negative & \cmark & \cmark \\ \hline

 \hline
\end{tabular}
 \caption{Definition of True Positive (TP), False Positive (FP), False Negative (FN) and True Negative (TN). \cmark means the noisy input token or prediction the same as the ground truth, and vice versa for \xmark. }
   \label{tab:metric_definition}
 \end{center}
\end{table}

We compare word-level precision, recall and $F_{0.5}$ score, which emphasizes precision more. We also provide accuracy for reference in Table \ref{tab:result}, because both of the baselines were evaluated with accuracy score. Table \ref{tab:metric_definition} shows the definition of true positive (TP), false positive (FP), false negative (FN) and true negative (TN) in this work to avoid confusions. We calculate them using the following equations:

{\small
\begin{equation*}
\begin{aligned}
accuracy &= (TP + TN) / (TP + FP + FN + TN) \\
precision &= TP / (TP + FP) \\
recall &= TP / (TP + FN) \\
F_\beta &= (1 + \beta^2) \cdot \dfrac{ precision \cdot recall } {(\beta^2 \cdot precision) + recall}
\end{aligned}
\end{equation*}
}

where $\beta=0.5$ in this work.

\paragraph{Baselines.} \citet{sakaguchi2017robsut} proposed semi-character recurrent neural network (ScRNN), which takes the first and the last character as well as the bag-of-word of the rest of the characters as features for each word. Then they used an LSTM \cite{hochreiter1997long} to predict each original word. \citet{wang2019learning} proposed MUDE, which uses a transformer-encoder \cite{vaswani2017attention} to encode character sequences as word representations and used an LSTM \cite{hochreiter1997long} for the correction of each word. They also used a Gated Recurrent Units (GRU) \cite{cho2014learning} to perform the character-level correction as an auxiliary task during training. We train ScRNN \cite{sakaguchi2017robsut} and MUDE \cite{wang2019learning}, both of which are \emph{stand-alone} neural spelling correctors, on our dataset as baselines. 

\paragraph{Overview.} As row \ref{model:ernie_randchar} of Table \ref{tab:result} shows, fine-tuning the Subword (WordPiece \cite{Peters:2018}) encoder model with LM initialization (ERNIE 2.0 \cite{sun2019ernie}) on the augmented dataset with synthetic character-level misspellings yields the best performance. Without leveraging a pre-trained LM, the Word+Char Encoder model trained on the augmented dataset with synthetic character-level misspellings performs the best (row \ref{model:word_char_randchar}). In fact, the differences between these approaches are small.

In Table \ref{tab:detailed_result}, we calculate real-word and non-word correction performance to explain the effect of each training technique applied. Note that as shown in Figure \ref{fig:spelling_corrector}, because non-word misspellings are pre-processed already, the detection of these non-word misspellings can be trivially accomplished, which results in all models having non-word recall of 1.000.

As Table \ref{tab:detailed_result} shows, strong models overall perform well on both real-word misspellings and non-word misspellings. Although our models perform better on non-word misspellings than real-word misspellings, the significant improvement of our models over the baselines comes from the real-word misspellings, due to the usage of the pre-trained LM. In the following paragraphs, we state our claims and support them with our experimental results.

\paragraph{Spelling correction requires both spelling and context information.}
As Table \ref{tab:detailed_result} shows, without the context information, the character encoder model (row \ref{model:char}) performs poorly on real-word misspellings. On the contrary, word encoder model (row \ref{model:word}) performs well on real-word misspellings, but poorly on non-word misspellings, due to the lack of the spelling information. The combined Word+Char encoder model (row \ref{model:word_char}) leverages both spelling and context information and thus improves nearly 40\% absolute $F_{0.5}$ in Table \ref{tab:result}. It even outperforms the LM intialized model (row \ref{model:ernie}). 
Both of the baseline models (row \ref{model:ScRNN} and \ref{model:MUDE}) perform poorly, because they perform spelling corrections upon character sequences, which disregards the semantics of the context, as their poor real-word performance in Table \ref{tab:detailed_result} row \ref{model:ScRNN} and \ref{model:MUDE} suggests. On the other hand, since subword embeddings essentially subsume character embedding, an additional character encoder does not improve the performance of the Subword encoder model (Table \ref{tab:result} row \ref{model:wordpiece_char}).

\paragraph{Pre-trained LM facilitates spelling correction.}
As row \ref{model:ernie} of Table \ref{tab:result} shows, fine-tuning the model with a pre-trained LM weight initialization improves both precision and recall score over the Subword encoder model (row \ref{model:wordpiece}).
The LM pre-training mainly improves real-word recall as Table \ref{tab:detailed_result} row \ref{model:ernie} suggests. Pre-trained LMs are trained with multiple unsupervised pre-training tasks on a much larger corpus than ours, which virtually expands the training task and the training set. 

Because most neural language models are trained on the subword level, we are not able to obtain a pre-trained LM initialized version of Word+Char encoder model (row \ref{model:word_char}). Nonetheless, we hypothesize that such a model will yield a very promising performance given sufficient training data and proper LM pre-training tasks.

\paragraph{Training on additional synthetic character-level noise improves model robustness.}
As row \ref{model:word_char_randchar}, \ref{model:wordpiece_randchar} and \ref{model:ernie_randchar} of Table \ref{tab:result} and \ref{tab:detailed_result} shows, in addition to frequently occurring natural misspellings, training models on the texts with synthetic character-level noise improves the test performance, which is mainly contributed by the improvement of precision on non-word misspellings. Note that the \emph{train} and \emph{dev} set only cover 80\% of the candidate natural misspellings. Adding character-level noise in the training data essentially increases the variety of the missplelling patterns, which makes the model more robust to unseen misspelling patterns.

\section{Related Work and Background}
Many approaches are proposed for spelling correction \cite{formiga2012dealing, kukich1992techniques, whitelaw2009using, zhang2006discriminative,flor2012four, carlson2007memory, flor2012using}, such as edit-distance based approaches \cite{damerau1964technique, levenshtein1966binary, bard2007spelling, kukich1992techniques, brill2000improved, de2013effective, pande2017effective}, approaches based on statistical machine translation \cite{chiu2013chinese, hasan2015spelling, liu2013hybrid}, and spelling correction for user search queries \cite{cucerzan2004spelling, gao2010large}.  Most of them do not use contextual information, and some use simple contextual features \cite{whitelaw2009using, flor2012four, carlson2007memory, flor2012using}.

In recent years, there are some attempts to develop better spelling correction algorithms based on neural nets \cite{etoori2018automatic}. Similar to our baselines ScRNN \cite{sakaguchi2017robsut} and MUDE \cite{wang2019learning}, \citet{li2018spelling} proposed a nested RNN to hierarchically encode characters to word representations, then correct each word using a nested GRU \cite{cho2014learning}. However, these previous works either only train models on natural misspellings \cite{sakaguchi2017robsut} or synthetic misspellings \cite{wang2019learning}, and only focus on denoising the input texts from orthographic perspective without leveraging the retained semantics of the noisy input.

On the other hand, Tal Weiss proposed Deep Spelling \cite{weissdeep}, which uses the sequence-to-sequence architecture \cite{sutskever2014sequence,bahdanau2014neural} to generate corrected sentences. Note that Deep Spelling is essentially not a spelling corrector since spelling correction must focus only on the misspelled words, not on transforming the whole sentences. For similar reasons, spelling correction is also different from GEC (Grammar Error Correction) \cite{zhang2014unified, junczys2018approaching}.

As a background, recently pre-trained neural LMs \cite{Peters:2018,devlin2018bert,yang2019xlnet,radford2019language,sun2019ernie} trained on large corpus on various pre-training tasks have made an enormous success on various benchmarks. These LMs captures the probability of a word or a sentence given their context, which plays a crucial role in correcting real-word misspellings. However, all of the LMs mentioned are based on subword embeddings, such as WordPiece \cite{Peters:2018} or Byte Pair Encoding \cite{gage1994new} to avoid OOV words. 

\section{Conclusion}
We leverage novel approaches to combine spelling and context information for \emph{stand-alone} spelling correction, and achieved state-of-the-art performance.
Our experiments gives insights on how to build a strong \emph{stand-alone} spelling corrector: (1) combine both spelling and context information, (2) leverage a pre-trained LM and (3) use the synthetic character-level noise.


\bibliography{anthology,emnlp2020}
\bibliographystyle{acl_natbib}

\clearpage
\appendix

\section{Appendices}
\label{sec:appendix}

\subsection{Dataset Details}
\label{sec:dataset_detail}
We keep the most frequent words in the $1$-Billion-Word-Language-Model-Benchmark dataset \cite{chelba2013one} as our word vocabulary $\Psi_w$, and all characters in $\Psi_w$ to form our character vocabulary $\Psi_c$. After deleting sentences containing OOV words, we randomly divide them into three datasets $S_{train}$, $S_{dev}$ and $S_{test}$. We merge the two word-level misspelling lists \cite{mitton1985corpora,belinkov2017synthetic} to get a misspelling list $\Omega$. We randomly choose $80\%$ of all misspellings in $\Omega$ to form a \emph{known-misspelling-list}, $\hat{\Omega}$. 

To strengthen the robustness of the model to various noisy spellings, we also utilize the methods in \citet{belinkov2017synthetic} , namely, \emph{swap}, \emph{middle random}, \emph{fully random} and \emph{keyboard type}, to generate character-level synthetic misspellings. To encourage the model to learn contextual information, we add an additional method, \emph{random generate}, to generate arbitrary character sequences as misspellings.

While replacing gold words with misspellings, for a sentence with $n$ words, the number of replaced words is $m=\max(\lfloor\alpha n\rfloor, 1)$,  where $\alpha=\min(|\mathcal{N}(0, 0.2)|, 1.0)$ and $\mathcal{N}$ represents a Gaussian distribution.

The \emph{dev} set is created with misspellings from sampled list $\hat{\Omega}$, and the \emph{test} set is created with misspellings from the full list $\Omega$. We compare 2 \emph{train} sets, the first has only natural misspellings from $\hat{\Omega}$, and the second has natural misspellings as well as synthetic misspellings, which is denoted as $+\mathit{random}$ $\mathit{char}$ in Section \ref{sec:results}. We always use the same \emph{dev} set and \emph{test} set that only contain natural misspellings for comparison.

Table \ref{tab:dataset} shows the parameters of our \emph{stand-alone} spelling correction dataset. We will release the dataset and codes after this paper is published.

\begin{table}[h]
\begin{center}
\begin{tabular}{c|c} 
 \hline
Parameter Name & Value \\ \hline
$|\Psi_w|$ & $50000$ \\ \hline
$|\Psi_c|$ & $130$ \\ \hline
$max\_sent\_len$ & $200$ \\ \hline
$max\_word\_len$ & $20$ \\ \hline
$|S_1|$ & $17971548$ \\ \hline
$|S_2|$ & $5985$ \\ \hline
$|S_3|$ & $5862$ \\ \hline
\end{tabular}
 \caption{Parameters of our \emph{stand-alone} spelling correction dataset.}
   \label{tab:dataset}
 \end{center}
\end{table}

\subsection{Implementation Details}

\begin{table}[h]
\begin{center}
\begin{tabular}{c|c|c|c} 
 \hline
Parameter Name & Word & Subword & Char \\ \hline
max seq length & 256 & 256 & 20 \\ \hline
hidden size & 512 & 768 & 256 \\ \hline
\# hidden layers & 6 & 12 & 4 \\ \hline
\# attention heads & 8 & 12 & 8 \\ \hline

 \hline
\end{tabular}
 \caption{Hyper-parameters of word encoders, Subword(WordPiece \cite{wu2016google}) encoders and character encoders.}
   \label{tab:parameters}
 \end{center}
\end{table}

We use PaddlePaddle \footnote{\url{https://github.com/PaddlePaddle/Paddle}} for the network implementation and keep the same configuration for the Subword encoders as ERNIE 2.0 \cite{sun2019ernie}. We tune the models by grid search on the \emph{dev} set according to $F_{0.5}$ score. The detailed hyper-parameters shown in Table \ref{tab:parameters}. In addition, we use Adam optimizer \cite{kingma2014adam} with learning rate of 5e-5 as well as linear decay. We used 10 GeForce GTX 1080 Ti or RTX 2080Ti to train each model until convergence, which takes a few days.

\end{document}